\documentclass[conference]{IEEEtran}
\IEEEoverridecommandlockouts
\usepackage{cite}
\usepackage{amsmath,amssymb,amsfonts}
\usepackage{algorithmic}
\usepackage{graphicx}
\usepackage{textcomp}
\usepackage{xcolor}
\usepackage{booktabs}
\def\BibTeX{{\rm B\kern-.05em{\sc i\kern-.025em b}\kern-.08em
    T\kern-.1667em\lower.7ex\hbox{E}\kern-.125emX}}
\begin{document}

\title{A Novel Convolution and Attention Mechanism-based Model for 6D Object Pose Estimation}

\author{
Alexander Du$^1$ \\
California Institute of Technology \\
\and
Xiujin Liu$^1$ \\
University of Michigan \\
}

\maketitle
\def\thefootnote{1}\footnotetext{These authors contributed equally to this work}

\begin{abstract}
This paper proposes PoseLecTr, a graph-based encoder-decoder framework that integrates a novel Legendre convolution with attention mechanisms for six-degree-of-freedom (6-DOF) object pose estimation from monocular RGB images. Conventional learning-based approaches predominantly rely on grid-structured convolutions, which can limit their ability to model higher-order and long-range dependencies among image features, especially in cluttered or occluded scenes. PoseLecTr addresses this limitation by constructing a graph representation from image features, where spatial relationships are explicitly modeled through graph connectivity. The proposed framework incorporates a Legendre convolution layer to improve numerical stability in graph convolution, together with spatial-attention and self-attention distillation to enhance feature selection.

Experiments conducted on the LINEMOD\cite{hinterstoisser2013model}, Occluded LINEMOD\cite{brachmann2014learning}, and YCB-VIDEO datasets\cite{xiang2017posecnn}
demonstrate that our method achieves competitive performance and shows consistent improvements across a wide range of objects and scene complexities.
\end{abstract}


\section{Introduction}
\label{sec:intro}

Recognizing objects and estimating their 6-DOF pose  $(x,y,z,\theta,\psi,\phi)$ in real-world environments is a fundamental challenge in computer vision and robotic perception, with broad applications in autonomous driving\cite{hoque2023deep}, augmented reality\cite{manawadu2024advancing}, and industrial automation\cite{govi2024addressing}. Traditional approaches which rely on matching
feature points and Perspective-n-Point (PnP) typically require clear textures, and lack effectiveness for texture-less objects. More recent methods incorporate deep learning to analyze large datasets and extract robust features from images. However, these approaches often struggle to capture dependencies among
extracted features, partly due to the inherent limitations of grid-based convolutional architectures. To address
these limitations, we propose PoseLecTr, a graph-based encoder-decoder
architecture that encodes image features as nodes and
utilizes an adjacency matrix to model their relationships. This
design leverages a graph neural network (GNN) to capture both local and global feature dependencies, while spatial-attention, self-attention
distillation, and a Legendre convolution method further
refine accuracy.

Our key contributions are as follows:

\paragraph{Graph-Based Representation} 
We introduce a novel graph-based
representation for image features, enabling more flexible
and expressive feature interactions compared to traditional
grid-based CNNs.

\paragraph{Legendre Convolution} 
We propose a novel Legendre convolution
layer, which significantly improves numerical stability
compared to standard graph convolutional layers.

\paragraph{Spatial-Attention and Self-Attention Distillation} 
We incorporate
advanced spatial-attention and self-attention distillation
mechanisms to enhance feature extraction and focus on
critical spatial information.

\section{Related Works}

Accurate 6D object pose estimation determines both the orientation and position of an object in three-dimensional space.
Non-learning approaches have relied on matching feature points
between 3D model projections and images\cite{lepetit2009ep, li2012robust, liu2025gnc}, or on offline
template databases for pose retrieval\cite{ulrich2011combining,payet2011contours}. The advancement of deep learning has significantly accelerated the progress of 6-DOF pose estimation, enabling accurate pose prediction from a single RGB image. Convolutional neural networks (CNNs)-based methods such as  \cite{xiang2017posecnn, peng2019pvnet, li2018deepim} show improved performance for 3D position and rotation estimation, but require large amounts of labeled data. More recent work\cite{wen2024foundationpose} eliminates the need for large amounts of labeled data and enables 6D pose estimation on novel objects. While grid-based convolutional models may overlook high-level feature dependencies, recent work has increasingly focused on graph-based neural networks\cite{yin2021graph, lian2023checkerpose, rezazadeh2023hierarchical}, which can capture complex relationships through node-grid representations and offer a promising avenue
for advancing 6D pose estimation.

Several challenges remain. Existing 6-DOF pose estimation methods still primarily rely on grid-based convolutional neural networks, which are effective for extracting features from images. However, traditional 2D convolution can only capture features from neighboring pixels and is limited in its ability to capture the correlation between different features at higher levels of abstraction. Graph-based representations offer a significant advantage due to their flexibility. Graphs can be constructed based on semantic similarity rather than just spatial proximity. Although graph-based neural networks have shown promise, several challenges remain in effectively applying graph-based models to 6-DOF pose estimation. In this work, we developed a graph-based encoder-decoder architecture with a novel Legendre convolutional layer with attention mechanisms to address the challenges above and to improve accuracy for pose estimation.
\section{Methodology}
\begin{figure*}[ht]
  \centering
  \includegraphics[scale=0.45]{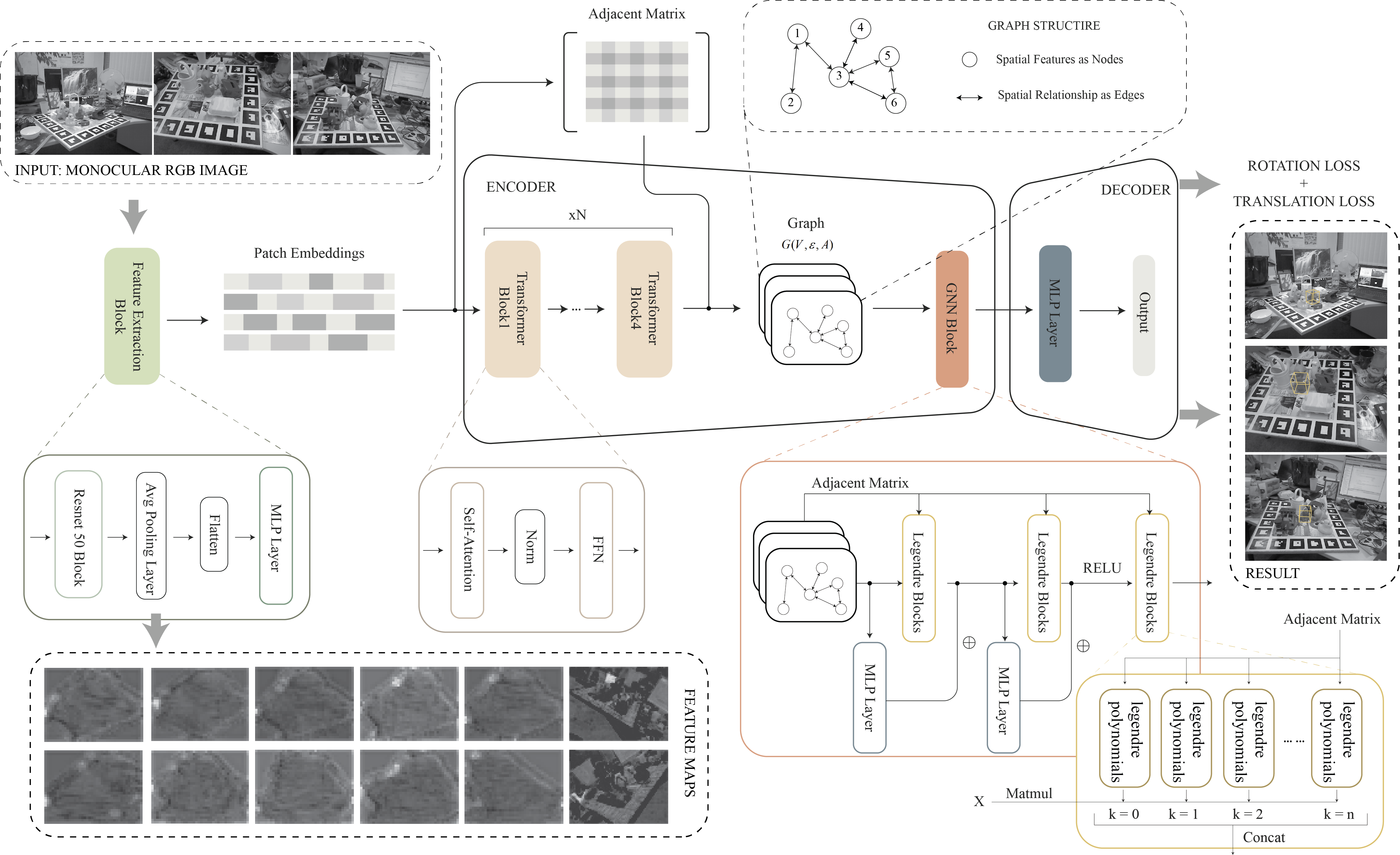}
  \caption{The architecture of PoseLecTr}
  \label{fig:vis}
\vspace{-0.6cm}
\end{figure*}
\subsection{Architecture}
Fig. \ref{fig:vis} depicts the structure of PoseLecTr. PoseLecTr has an encoder-decoder structure. Before the encoder, ResNet-50\cite{he2016deep} is used as the backbone to extract image features. The encoder, which serves as a feature aggregation neck, consists of spatial-attention blocks to capture complex spatial dependencies on 3D objects, and a graph neural network block with Legendre convolution blocks. The decoder layer outputs the final 6D pose estimation.
\subsection{Input and Output}
The goal of PoseLecTr is to estimate the homogeneous transformation matrix of camera pose related to object pose. The
inputs for the PoseLecTr are monocular RGB images of size H×W×3, where H and W are the height and width of the images. The images contain objects with specific camera poses. The output of the PoseLecTr is a predicted 12×1 vector, where the first 9 elements are reshaped to a 3×3 rotation matrix, and the remaining 3 elements represent the translation vector.
\subsection{Graph Neural Network}
A graph structure is used to organize spatial features of 3D objects extracted from images. Formally, the graph is denoted as $G(V, E, A)$, where $V=\{v_1,v_2,...,v_N\}$ denotes a set of nodes $N$ corresponding to feature maps of extraction layers; $E \subseteq V \times V$ is a set of edges representing the connections between nodes; and $A \in \mathbb{R}^{N \times N}$ is a real-number adjacency matrix constructed based on the four-neighbors of feature maps, capturing the relational information between different regions of the images.
\subsection{Legendre Convolution Layer} 
Legendre polynomials offer the following advantageous mathematical properties for functional representation.\cite{mccarthy1993generalized}

\paragraph{Orthogonality}
Polynomials of different orders are mutually orthogonal under the standard inner product on $\left[-1,1\right]$, allowing expansion coefficients to be computed independently and simplifying the fitting process. 
\paragraph{Numerical Stability}
They exhibit strong numerical stability, which helps suppress oscillatory artifacts such as Runge’s phenomenon\cite{press2007numerical} when using high-order approximations. 
\paragraph{Recursion}
Legendre polynomials satisfy a simple recursive formulation, enabling efficient construction of higher-order terms. 

For these reasons, we integrate Legendre polynomials into our proposed Legendre Convolution Layer to achieve stable, efficient, and mathematically well-structured feature modeling.

The Legendre convolution layer includes the spatial graph Laplacian embedding for encoding the spatial relationships within the feature maps. To characterize the interconnections between nodes in the graph, we utilize the regularized Laplacian matrix denoted as $L=I-D^{-0.5}AD^{-0.5}$, where $A$ represents the adjacency matrix, $I$ is the identity matrix, and $D$ is the matrix of node degrees in diagonal form. As $L$ is a symmetric positive semi-definite matrix, its eigenvectors are orthogonal in pairs, and the eigenvalues are distinct and non-negative. Therefore, $L$ can be simplified to the form of $L=U$$\Delta$$U^{T}$, where $U$ is a matrix composed of eigenvectors and $\delta$ is a matrix composed of eigenvalues. In our model, the output of Legendre convolution layer is defined as 
$Y_{output}=$$\sigma$$(Ug_{\theta}(\Delta)U^{T}X_{input})$, where $\sigma()$ denotes the activation function, and $\theta$ is the parameter value for iterative updates. Here, $g_{\theta}(\Delta)$ signifies the convolution kernel operating on a graph.

Legendre polynomials can be written recursively as
\begin{equation}
\begin{aligned}
    &P_0=1\\
    &P_1(x)=x(n+1)\\
    &p_{n+1}(x)=(2n+1)xP_n(x)-nP_{n-1}(x)\\
    &n=1,2,3...
    \label{equ:lp}
\end{aligned}
\end{equation}
In this work, Legendre polynomials are used to approximate the convolution kernel. Legendre polynomials are defined as 
\begin{equation}
\begin{aligned}
    &P_0(x)=1\\
    &P_n(x)=\frac{1}{2^nn}\frac{d^n}{dx^n}[(x^2-1)^n]\\
    &n=1,2,3...
    \label{equ:legendre}
\end{aligned}
\end{equation}
We write the Legendre convolution block as
\begin{equation}
\begin{aligned} 
&g_\theta(\Delta)=Concat(Matmul(\alpha_iP_i(\widetilde{\Delta}), x))\\
    &i = 0, 1, ..., k
\label{equ:legendre_block}
\end{aligned} 
\end{equation}
 
\subsection{Spatial Feature Self-Attention}
A spatial feature self-attention layer is used to capture spatial features of 3D objects from monocular RGB images for more accurate 6-DOF object pose estimation. Let $\mathbb{X}=(X_1,X_2,...,X_T)\subset\mathbb{R}^{T*N*D}$ denote the input images, where $T$ represents number of images, $N$ represents number of features extracted from one image, and $D$ represents the feature dimension. The query, key, and value matrices of self-attention operations can be represented as:
\begin{equation}
q^{(i)}=W_qx^{(i)}, k^{(i)}=W_kx^{(i)}, v^{(i)}=W_vx^{(i)} for i \in [1, T],
\end{equation}
where $W_q, W_k, W_v$ $\in$ $R^{d*d'}$ are learnable parameters and $d'$ is the dimension of the query, key, and value matrices. The self-attention mechanism, $SA$, can be expressed as\cite{vaswani2017attention}:
\begin{equation}
SA(q,k,v)=Softmax(\frac{qk^T}{\sqrt d})v.
    \label{equ:att}
\end{equation}

The spatial dependencies between nodes are different in different $X_i$. The key to the attention mechanism is the mapping function, which encodes the relative importance of the input elements, mapping the values to probabilities. The commonly used softmax function generates dense attention. Note that in the task of 6-DOF pose estimation, the distribution of feature relevance can be highly imbalanced across different regions of the image. Therefore, we introduce a new spatial feature attention module, written as 
\begin{equation}
    SFA(q,k,v)=Sparsemax(\frac{qk^T}{\sqrt d})v.
    \label{equ:att2}
\end{equation}
In contrast to the Softmax function in Equation~\ref{equ:att}, the Sparsemax function generates sparse probability distributions. Sparsemax has the ability to selectively emphasize the most significant features while disregarding less relevant ones, thus offering clearer decision boundaries \cite{martins2016softmax}.  Furthermore, when handling complex image data for 6-DOF pose estimation, Sparsemax is better at handling the long-tail effects, ensuring the model focuses on the most critical spatial features.

\subsection{Loss Function}
In PoseLecTr, the total loss function consists of two components, rotation matrix loss and translation vector loss. These two losses evaluate  both the orientation and translation of the predicted poses relative to the ground truth. The first term of our loss function measures the discrepancy between the predicted rotation matrix $\hat{R}$ and the ground-truth rotation matrix $R$. We use the Frobenius norm of the difference between these two matrices, which  captures the magnitude differences between them. The loss function is defined as 
\begin{equation}
    L_R = \frac{1}{n}\sum_{i=1}^{n} \left|\left|\hat{R_i} - R_i\right|\right|_{F}^{2},
    \label{equ:loss1}
\end{equation}
where $\left|\left| \cdot \right|\right|_{F}$ denotes the Frobenius norm. 

The second term is the Mean Squared Error (MSE) between the predicted translation vector $\hat{t}$ and the ground truth translation vector $t$. The translation loss function is given by
\begin{equation}
    L_t = \frac{1}{n}\sum_{i=1}^{n} \left|\left|\hat{t_i} - t_i\right|\right|_{F}^{2},
    \label{equ:loss2}
\end{equation}
where $\hat{t_i}$ represents the predicted translation vector and $t_i$ is the corresponding ground truth translation vector.

The final loss function is a weighted sum of the rotation loss function and translation loss function. The total loss function is defined as
\begin{equation}
    L = \eta_1 L_R + \eta_2 L_t,
\label{equ:lossall}
\end{equation}
where $\eta_1$ and $\eta_2$ are weighting parameters for each loss term. 
\section{Experiments}

\begin{table*}[t]
\centering
\caption{Quantitative evaluation of different 6D pose estimation baselines on LINEMOD dataset.}
\small
\setlength{\tabcolsep}{4.5pt}
\renewcommand{\arraystretch}{1.1}

\resizebox{\textwidth}{!}{
\begin{tabular}{lcccccccccccccccccccc}
\toprule
& \multicolumn{2}{c}{\textbf{PoseCNN \cite{xiang2017posecnn}}}
& \multicolumn{2}{c}{\textbf{DenseFusion \cite{wang2019densefusion}}}
& \multicolumn{2}{c}{\textbf{PVNet \cite{peng2019pvnet}}}
& \multicolumn{2}{c}{\textbf{PoseRBPF \cite{deng2021poserbpf}}}
& \multicolumn{2}{c}{\textbf{DeepIM \cite{li2018deepim}}}
& \multicolumn{2}{c}{\textbf{GC-Pose \cite{zhao2023learning}}}
& \multicolumn{2}{c}{\textbf{G2LNet \cite{chen2020g2l}}}
& \multicolumn{2}{c}{\textbf{TransPose \cite{lin2024transpose}}}
& \multicolumn{2}{c}{\textbf{Ours}} \\

Object
& ADDS & ADD
& ADDS & ADD
& ADDS & ADD
& ADDS & ADD
& ADDS & ADD
& ADDS & ADD
& ADDS & ADD
& ADDS & ADD
& ADDS & ADD
\\
\midrule                      
ape & 90.4 & 90.3 & 96.6 & 95.7 & 92.4 & 91.3 & 89.1 & 88.4 & 91.0 & 90.8 & 93.4 & 92.8 &95.2 & 94.5 & 96.0 & 95.8 & 98.9 & 98.0\\
benchvise & 89.8 & 89.6 & 98.0 & 97.7 & 89.2 & 88.6 & 86.7 & 86.0 & 90.8 & 90.4 & 94.1 & 93.0 &93.4 & 92.9 & 96.6 & 96.4 & 99.6 & 99.4\\

cam & 96.6 & 96.2 & 99.8 & 98.7 & 92.6 & 91.7 & 91.7 & 91.6 & 94.3 & 93.2 & 94.4 & 94.1 &97.3 & 96.6 & 97.3 & 96.3 & 98.4 & 98.3\\
                       
can & 89.8 & 89.7 & 96.1 & 95.4 & 90.8 & 90.1 & 89.3 & 88.2 & 93.4 & 92.3 & 92.5 & 92.4 & 94.4 & 94.0 & 96.4 & 95.6 & 100 & 99.9\\
                       
cat & 91.0 & 89.9 & 97.8 & 97.1 & 89.9 & 89.7 & 85.7 & 85.3 & 91.1 & 91.1 & 93.4 & 92.7
& 94.6 & 93.5 & 96.7 & 96.6 & 99.4 & 98.8\\
                       
driller & 96.9 & 96.2 & 100 & 99.2 & 92.7 & 91.6 & 92.2 & 91.6 & 95.5 & 94.3 & 96.0 & 95.4
& 96.9 & 95.9 & 97.8 & 97.0 & 99.1 & 98.8\\

duck & 91.5 & 90.9 & 96.0 & 95.6 & 90.5 & 89.9 & 88.3 & 88.1 & 92.4 & 92.1 & 93.3 & 92.2
& 94.3 & 94.1 & 95.4 & 94.9 & 100 & 99.3\\

eggbox & 94.5 & 93.9 & 96.9 & 96.8 & 92.0 & 91.4 & 90.2 & 90.0 & 93.1 & 92.1 & 95.5 & 94.3
& 96.1 & 95.5 & 98.1 & 97.3 & 99.5 & 98.9\\

glue & 79.2 & 78.5 & 87.8 & 87.6 & 80.8 & 80.2 & 79.4 & 78.7 & 83.5 & 82.5& 85.8 & 85.1
& 86.6 & 85.5 & 88.8 & 87.7 & 91.3 & 92.7\\

holepuncher & 82.2 & 82.1 & 100 & 99.6 & 85.6 & 85.6 & 83.5 & 82.7 & 85.2 & 85.1 & 87.4 & 86.9
& 89.1 & 88.4 & 91.0 & 90.1 & 94.7 & 93.7\\

iron & 91.7 & 90.9 & 100 & 99.3 & 95.0 & 94.1 & 91.1 & 91.0 & 94.2 & 93.4 & 95.9 & 95.5
& 97.8 & 97.1 & 98.4 & 98.3 & 99.3 & 99.2\\

lamp & 82.8 & 82.2 & 79.4 & 78.2 & 83.1 & 82.0 & 80.5 & 79.6 & 83.6 & 82.6 & 85.9 & 85.3
& 86.6 & 85.6 & 88.6 & 87.8 & 91.1 & 91.1\\

phone & 83.4 & 83.3 & 96.7 & 96.5 & 86.0 & 85.2 & 83.8 & 83.5 & 87.6 & 87.1 & 90.1 & 89.2
& 90.9 & 90.5 & 93.1 & 92.2 & 99.6 & 98.6\\
\midrule

\textbf{MEAN} & 89.2 & 88.7 & 95.8 & 95.2 & 89.3 & 88.6 & 87.0 & 86.5 
& 90.4 & 89.8 & 92.1 & 91.5 & 93.3 & 92.6 & 94.9 & 94.3 
& 97.8 & 97.4\\
\bottomrule
\end{tabular}}
\label{tab:linemode}
\end{table*}

\begin{table*}[t]
\centering
\caption{Quantitative evaluation of different 6D pose estimation baselines on YCB-Video dataset.}
\small
\setlength{\tabcolsep}{4.5pt}   
\renewcommand{\arraystretch}{1.1}   

\resizebox{\textwidth}{!}{
\begin{tabular}{lcccccccccccccccccccc}
\toprule

& \multicolumn{2}{c}{\textbf{PoseCNN \cite{xiang2017posecnn}}}
& \multicolumn{2}{c}{\textbf{DenseFusion \cite{wang2019densefusion}}}
& \multicolumn{2}{c}{\textbf{PVNet \cite{peng2019pvnet}}}
& \multicolumn{2}{c}{\textbf{PoseRBPF \cite{deng2021poserbpf}}}
& \multicolumn{2}{c}{\textbf{DeepIM \cite{li2018deepim}}}
& \multicolumn{2}{c}{\textbf{GC-Pose \cite{zhao2023learning}}}
& \multicolumn{2}{c}{\textbf{G2LNet \cite{chen2020g2l}}}
& \multicolumn{2}{c}{\textbf{TransPose \cite{lin2024transpose}}}
& \multicolumn{2}{c}{\textbf{Ours}} \\
Object
& ADDS & ADD
& ADDS & ADD
& ADDS & ADD
& ADDS & ADD
& ADDS & ADD
& ADDS & ADD
& ADDS & ADD
& ADDS & ADD
& ADDS & ADD
\\
\midrule
                       
002\_master\_chef\_can & 89.0 & 85.3 & 93.9 & 92.2 & 89.3 & 87.1 & 86.5 & 84.4 & 90.2 & 89.6 & 91.5 & 89.6 & 92.7 & 89.0 & 94.0 & 93.7 & 96.4 & 93.9\\

003\_cracker\_box & 88.5 & 86.6 & 96.5 & 94.5 & 87.8 & 85.2 & 85.0 & 83.6 & 89.9 & 87.2 & 91.0 & 88.2
& 92.5 & 89.1 & 94.7 & 93.2 & 97.8 & 97.1\\

004\_sugar\_box & 94.3 & 91.9 & 97.6 & 95.7 & 91.2 & 87.3 & 89.7 & 87.8 & 92.3 & 91.2 & 93.5 & 92.0
& 94.8 & 91.7 & 96.0 & 95.8 & 97.5 & 95.5\\
                       
005\_tomato\_soup\_can & 89.1 & 87.3 & 95.0 & 94.5 & 88.9 & 87.8 & 86.8 & 83.6 & 90.7 & 89.9 & 91.9 & 90.1
& 93.2 & 90.0 & 94.5 & 93.2 & 98.5 & 96.7\\
                       
006\_mustard\_bottle & 92.0 & 91.6 & 95.8 & 91.9 & 90.5 & 89.2 & 88.6 & 86.1 & 91.8 & 90.3 & 93.0 & 92.3
& 94.2 & 92.6 & 95.5 & 95.0 & 97.3 & 96.3\\
                       
007\_tuna\_fish\_can & 78.0 & 77.1 & 86.5 & 83.0 & 79.3 & 75.7 & 77.5 & 75.2 & 81.8 & 78.6 & 83.2 & 81.0
& 84.5 & 82.7 & 85.8 & 85.6 & 91.2 & 88.0\\

008\_pudding\_box & 81.4 & 77.5 & 97.9 & 96.6 & 83.7 & 83.2 & 80.9 & 79.3 & 84.2 & 81.7 & 85.6 & 85.1
& 87.1 & 85.2 & 89.3 & 87.3 & 91.8 & 89.8\\

009\_gelatin\_box & 90.4 & 90.0 & 97.8 & 94.9 & 92.1 & 88.6 & 89.8 & 89.7 & 93.0 & 91.0 & 94.2 & 93.0
& 95.6 & 93.2 & 96.7 & 93.7 & 98.7 & 95.8\\

010\_potted\_meat\_can & 81.7 & 77.9 & 77.8 & 74.4 & 80.3 & 78.9 & 78.2 & 76.9 & 82.1 & 79.9 & 83.5 & 80.2
& 84.9 & 83.5 & 86.3 & 82.6 & 90.2 & 88.2\\

011\_banana & 82.6 & 80.1 & 94.9 & 93.5 & 84.5 & 81.1 & 83.2 & 79.6 & 86.5 & 84.5 & 87.8 & 84.4
& 89.1 & 88.2 & 91.3 & 88.1 & 96.9 & 95.2\\
\midrule

\textbf{MEAN} & 86.7 & 84.5 & 93.4 & 91.1 & 86.7 & 84.4 & 84.6 & 82.6 & 88.3 & 86.4 & 89.5 & 87.6
& 90.1 & 88.5 & 92.4 & 90.8 & 95.6 & 93.6\\
\bottomrule
\end{tabular}}
\label{tab:ycb}
\end{table*}

\begin{table*}[t]
\centering
\caption{Quantitative evaluation of different 6D pose estimation baselines on Occluded LINEMOD dataset.}
\small
\setlength{\tabcolsep}{4.5pt}   
\renewcommand{\arraystretch}{1.1}   

\resizebox{\textwidth}{!}{
\begin{tabular}{lcccccccccccccccccc}
\toprule

& \multicolumn{2}{c}{\textbf{PoseCNN \cite{xiang2017posecnn}}}
& \multicolumn{2}{c}{\textbf{DenseFusion \cite{wang2019densefusion}}}
& \multicolumn{2}{c}{\textbf{PVNet \cite{peng2019pvnet}}}
& \multicolumn{2}{c}{\textbf{PoseRBPF \cite{deng2021poserbpf}}}
& \multicolumn{2}{c}{\textbf{DeepIM \cite{li2018deepim}}}
& \multicolumn{2}{c}{\textbf{GC-Pose \cite{zhao2023learning}}}
& \multicolumn{2}{c}{\textbf{G2LNet \cite{chen2020g2l}}}
& \multicolumn{2}{c}{\textbf{TransPose \cite{lin2024transpose}}}
& \multicolumn{2}{c}{\textbf{Ours}} \\

Object
& ADDS & ADD
& ADDS & ADD
& ADDS & ADD
& ADDS & ADD
& ADDS & ADD
& ADDS & ADD
& ADDS & ADD
& ADDS & ADD
& ADDS & ADD
\\
\midrule
                       
ape & 86.8 & 85.5 & 91.6 & 90.0 & 89.0 & 87.9 & 84.1 & 82.9 & 88.1 & 86.4 & 90.6 & 89.5
& 91.2 & 89.8 & 93.1 & 91.8 & 93.7 & 92.3\\

can & 85.5 & 83.9 & 93.2 & 91.9 & 86.2 & 85.0 & 83.3 & 82.1 & 89.1 & 87.8 & 89.8 & 88.5
& 91.8 & 90.5 & 92.3 & 90.6 & 95.5 & 94.3\\

cat & 91.6 & 89.9 & 96.1 & 94.7 & 90.5 & 88.9 & 89.3 & 88.0 & 90.4 & 89.4 & 91.9 & 90.2
& 91.7 & 90.1 & 94.4 & 92.9 & 95.5 & 93.7\\
                       
driller & 87.3 & 85.6 & 94.6 & 93.3 & 86.5 & 84.9 & 86.2 & 84.6 & 89.5 & 87.9 & 90.7 & 89.2
& 91.8 & 90.2 & 94.2 & 92.6 & 94.3 & 92.9\\
                       
duck & 86.2 & 84.4 & 93.9 & 92.3 & 87.3 & 85.6 & 83.6 & 82.1 & 89.8 & 88.6 & 90.3 & 88.9
& 90.4 & 89.1 & 92.8 & 91.5 & 93.8 & 92.5\\
                       
eggbox & 93.2 & 92.2 & 96.9 & 95.4 & 89.0 & 87.9 & 88.5 & 87.3 & 90.8 & 89.1 & 90.9 & 89.2
& 92.8 & 91.8 & 93.6 & 92.4 & 95.2 & 93.6\\

glue & 87.6 & 86.4 & 92.6 & 91.0 & 87.9 & 86.5 & 83.7 & 82.6 & 88.7 & 87.1 & 90.6 & 89.4
& 91.5 & 89.9 & 92.5 & 91.5 & 92.8 & 91.4\\

holepuncher & 91.0 & 89.8 & 93.6 & 92.6 & 89.2 & 87.9 & 87.6 & 86.4 & 90.4 & 89.2 & 91.5 & 90.3
& 92.6 & 90.8 & 93.5 & 92.3 & 95.0 & 93.8\\
\midrule

\textbf{MEAN} & 88.7 & 87.2 & 94.1 & 92.7 & 88.2 & 86.8 & 85.8 & 84.5 & 89.6 & 88.2 & 90.8 & 89.4 & 91.7 & 90.3 & 93.3 & 92.0 & 94.5 & 93.1 \\
\bottomrule
\end{tabular}}
\label{tab:olinemod}
\end{table*}

\subsection{Datasets}
We evaluate our method on three widely recognized benchmarks: LINEMOD dataset \cite{hinterstoisser2013model}, YCB-Video dataset \cite{xiang2017posecnn} and Occluded LINEMOD dataset\cite{brachmann2014learning}. The LINEMOD dataset  consists of 15 registered video sequences, each containing over 1100 frames. These sequences feature 15 different texture-less household objects, captured in varying poses and lighting conditions. The YCB-Video dataset contains 92 real-world video sequences featuring 21 different objects from the YCB object set, including items such as power drills, cups, and bananas. The Occluded LINEMOD dataset introduces significant occlusion and clutter, facilitating robust testing of algorithm resilience under partial visibility.

\subsection{Comparison Methods}
We compared our PoseLecTr to the following eight baseline models that fall into three categories:

\paragraph{CNN-Based Models} PoseCNN \cite{xiang2017posecnn} and DenseFusion \cite{wang2019densefusion} are benchmark baselines for integrating RGB and depth data to achieve accurate 6-DOF pose estimations. PoseCNN focuses on predicting 3D translation and rotation directly from RGB images, while DenseFusion fuses dense pixel-wise information from both RGB and depth maps. PVNet \cite{peng2019pvnet} approaches pose estimation by regressing 2D keypoints from the image, followed by a PnP algorithm to compute the pose, ensuring precision through keypoint localization. PoseRBPF \cite{deng2021poserbpf} combines recurrent neural networks and particle filters to handle temporal dynamics in pose estimation, making it robust against occlusions and other temporal inconsistencies. DeepIM \cite{li2018deepim} iteratively refines an initial pose estimate by aligning rendered and observed images, providing a feedback loop that improves the pose estimation accuracy over several iterations.

\paragraph{GNN-Based Models} GC-Pose \cite{zhao2023learning} utilizes graph convolutional networks (GCN) to model spatial relationships between object parts as a graph, extracting features from RGB images to capture the geometric structure of the object. G2LNet \cite{chen2020g2l} Employs a GNN to learn spatial relationships between keypoints, using these to predict the 6-DOF pose of the object.

\paragraph{Self-Attention-Based Models} TransPose\cite{lin2024transpose} adopts a transformer-based architecture to model long-range dependencies in image data, leveraging self-attention mechanisms to focus on image regions most relevant to 6D pose prediction. 

\subsection{Implementation Details}
Our proposed methods are trained using the Adam optimizer, with an initial learning rate of $1e-4$, which is reduced by half after each epoch. The training process spans 8 epochs, incorporating early stopping to prevent overfitting. All baseline methods are configured according to their recommended settings, using a batch size of 32.

All experiments were carried out with an NVIDIA GeForce 4080 GPU and 64GB of RAM. The code was implemented with Ubuntu 18.04, PyTorch 1.16.10 and Python 3.9.0.

\subsection{Evaluation Metrics}
Two metrics are used to measure pose prediction performance, the Average Distance Deviation (ADD) and Average Distance Deviation with Symmetry (ADD-S), written as
\begin{equation}
    ADD=\frac{1}{m}\sum_{x\in M} \left|\left| (Rx+t)-(\hat{R}x+\hat{t}) \right|\right|.
    \label{equ:add}
\end{equation}
\begin{equation}
    ADD_S=\frac{1}{m}\sum_{x_1\in M} \min_{x_2\in M}\left|\left| (Rx_1+t)-(\hat{R}x_2+\hat{t}) \right|\right|.
   \label{equ:adds}
\end{equation}

The ADD metric quantifies the accuracy of the predicted attitude
by measuring the mean Euclidean distance between the corresponding
three-dimensional points of the model. Similarly, for symmetric objects, instead of calculating point-to-point distance, ADD-S measures the average minimum distance between each transformation point in the predicted attitude and the nearest point on the model transformed by the ground-truth pose. If ADD/ADD-S is less than 0.1*d, where d is the maximum measurement of the object, the posture is considered correct. 

\subsection{Quantitative Results}
Table \ref{tab:linemode}, Table \ref{tab:ycb} and Table \ref{tab:olinemod} show the experimental results on the LINEMOD dataset\cite{hinterstoisser2013model}, YCB-Video dataset\cite{xiang2017posecnn}, and Occluded LINEMOD dataset\cite{brachmann2014learning} respectively.

Specifically, for the YCB-Video dataset, on the object \textit{002\_master\_chef\_can}, PoseLecTr achieves the highest accuracy among the evaluated methods, with 96.4\% under ADD-S and 93.9\% under ADD metrics. On the object \textit{003\_cracker\_box}, PoseLecTr records the highest accuracies of 97.8\% and
97.1\%, while DenseFusion follows with 96.5\% and 94.5\%. More
challenging objects, such as \textit{007\_tuna\_fish\_can}, also highlight
PoseLecTr’s advantages, with accuracies of 91.2\% and
88\%. Considering the overall performance across all
frames, PoseLecTr achieves the highest mean accuracy in this comparison for both symmetric and asymmetric objects, scoring 95.6\% and 93.6\%, indicating consistently
strong performance across diverse object categories and evaluation settings.

On the LINEMOD dataset, PoseLecTr achieves mean
accuracies of 97.8\% and 97.4\% under the ADD-S and ADD
metrics, respectively, representing the highest results among the evaluated methods. 

On the more challenging Occluded LINEMOD dataset, PoseLecTr maintains its robustness by posting mean accuracies of 94.5\% and
93.1\%.
\subsection{Ablation Studies}
\begin{figure}[t]
  \centering
  \includegraphics[scale=0.55]{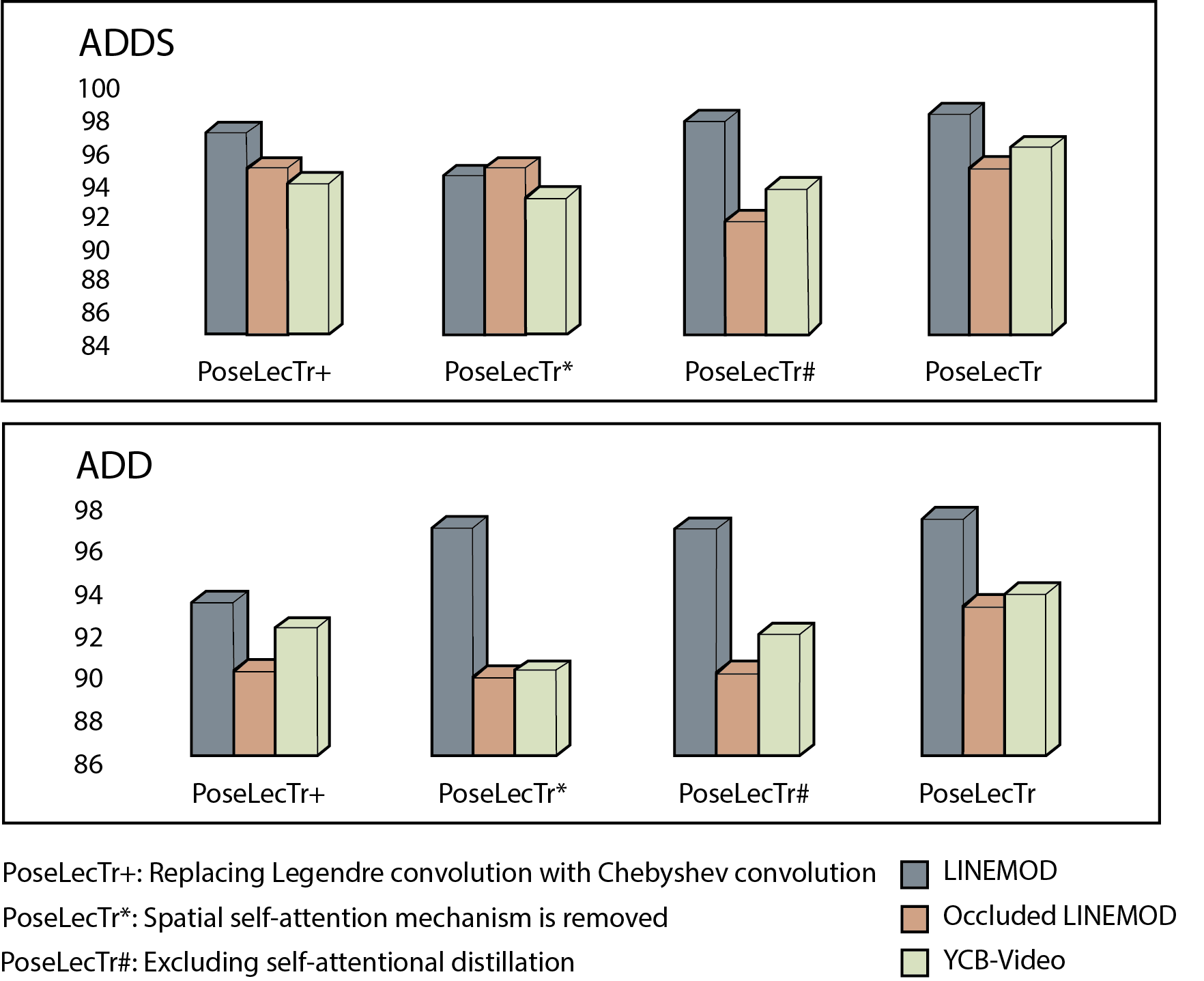}
  \caption{Visualization of Ablation Studies}
  \label{fig:abs}
\vspace{-0.6cm}
\end{figure}
As shown in Fig. \ref{fig:abs}, to further investigate the efficacy of PoseLecTr’s components, we conducted a comparative analysis using several modifications
of PoseLecTr.

From the results, we can derive the following insights:
\paragraph{Effect of Graph Convolution Type (PoseLecTr+)} Replacing
Legendre convolution with Chebyshev convolution
in PoseLecTr+ results in slightly lower performance across
all datasets. On the LINEMOD dataset, accuracy decreases
from 99.7\% (PoseLecTr) to 96.4\%, with similar declines on
Occluded LINEMOD (94.34\% vs. 95.54\%) and YCB-Video
(91.56\% vs. 93.36\%). These results indicate that Legendre convolution provides more effective modeling of both local and global contexts than Chebyshev convolution.
\paragraph{Effects of Spatial Feature Self-Attention (PoseLecTr*)}
When the spatial self-attention mechanism is removed
from PoseLecTr*, the accuracy on Occluded LINEMOD
(92.54\%) and YCB-Video (89.96\%) decreases more substantially.
This indicates that spatial self-attention is particularly
important in handling occlusion and complex
scenes. Meanwhile, the performance on LINEMOD (96.9\%)
remains relatively stable, suggesting that removing spatial
self-attention chiefly affects more challenging datasets
where spatial dependencies play a more critical role.
\paragraph{Effects of Self-Attentional Distillation (PoseLecTr\#)}
Excluding self-attentional distillation from PoseLecTr\# modules
leads to a decline in performance, particularly on
LINEMOD (92.64\%) and YCB-Video (89.56\%). These results
indicate that self-focused distillation is essential for
handling larger graph sizes and extended data. Without this
component, the model’s capacity to manage extensive spatial
relationships is diminished, thereby reducing its ability to
generalize to more complex environments.

\section{Conclusion}
In this paper, we present PoseLecTr, a novel graph-based
framework for 6D object pose estimation from monocular RGB images. By
incorporating Legendre convolution and attention mechanisms,
PoseLecTr advances accuracy in pose estimation
tasks. The introduction of Legendre convolution, with its
orthogonality and numerical stability properties, streamlines the
convolution process and enables more efficient feature extraction
compared to traditional convolution methods. Additionally, the
spatial feature self-attention and self-attention distillation modules
enhance the model’s ability to focus on critical spatial
features, thus improving the robustness and accuracy of pose estimation.
Experimental results on benchmark datasets such as LINEMOD\cite{hinterstoisser2013model},
Occluded LINEMOD\cite{brachmann2014learning} and YCB-Video\cite{xiang2017posecnn} demonstrate that PoseLecTr achieves competitive performance relative to existing state-of-the-art models,
particularly in complex scenes involving textureless, symmetric, or
highly occluded objects.

\bibliographystyle{IEEEbib}
\bibliography{icme2026references}

@article{hoque2023deep,
  title={Deep learning for 6D pose estimation of objects—A case study for autonomous driving},
  author={Hoque, Sabera and Xu, Shuxiang and Maiti, Ananda and Wei, Yuchen and Arafat, Md Yasir},
  journal={Expert Systems with Applications},
  volume={223},
  pages={119838},
  year={2023},
  publisher={Elsevier}
}

@article{manawadu2024advancing,
  title={Advancing 6D Pose Estimation in Augmented Reality--Overcoming Projection Ambiguity with Uncontrolled Imagery},
  author={Manawadu, Mayura and Park, Sieun and Park, Soon-Yong},
  journal={arXiv preprint arXiv:2403.13434},
  year={2024}
}

@article{govi2024addressing,
  title={Addressing challenges in industrial pick and place: A deep learning-based 6 Degrees-of-Freedom pose estimation solution},
  author={Govi, Elena and Sapienza, Davide and Toscani, Samuele and Cotti, Ivan and Franchini, Giorgia and Bertogna, Marko},
  journal={Computers in Industry},
  volume={161},
  pages={104130},
  year={2024},
  publisher={Elsevier}
}

@inproceedings{hinterstoisser2013model,
  title={Model based training, detection and pose estimation of texture-less 3d objects in heavily cluttered scenes},
  author={Hinterstoisser, Stefan and Lepetit, Vincent and Ilic, Slobodan and Holzer, Stefan and Bradski, Gary and Konolige, Kurt and Navab, Nassir},
  booktitle={Computer Vision--ACCV 2012: 11th Asian Conference on Computer Vision, Daejeon, Korea, November 5-9, 2012, Revised Selected Papers, Part I 11},
  pages={548--562},
  year={2013},
  organization={Springer}
}

@article{xiang2017posecnn,
  title={Posecnn: A convolutional neural network for 6d object pose estimation in cluttered scenes},
  author={Xiang, Yu and Schmidt, Tanner and Narayanan, Venkatraman and Fox, Dieter},
  journal={arXiv preprint arXiv:1711.00199},
  year={2017}
}

@inproceedings{brachmann2014learning,
  title={Learning 6d object pose estimation using 3d object coordinates},
  author={Brachmann, Eric and Krull, Alexander and Michel, Frank and Gumhold, Stefan and Shotton, Jamie and Rother, Carsten},
  booktitle={European conference on computer vision},
  pages={536--551},
  year={2014},
  organization={Springer}
}

@inproceedings{wang2019densefusion,
  title={Densefusion: 6d object pose estimation by iterative dense fusion},
  author={Wang, Chen and Xu, Danfei and Zhu, Yuke and Mart{\'\i}n-Mart{\'\i}n, Roberto and Lu, Cewu and Fei-Fei, Li and Savarese, Silvio},
  booktitle={Proceedings of the IEEE/CVF conference on computer vision and pattern recognition},
  pages={3343--3352},
  year={2019}
}

@inproceedings{peng2019pvnet,
  title={Pvnet: Pixel-wise voting network for 6dof pose estimation},
  author={Peng, Sida and Liu, Yuan and Huang, Qixing and Zhou, Xiaowei and Bao, Hujun},
  booktitle={Proceedings of the IEEE/CVF conference on computer vision and pattern recognition},
  pages={4561--4570},
  year={2019}
}

@article{deng2021poserbpf,
  title={PoseRBPF: A Rao--Blackwellized particle filter for 6-D object pose tracking},
  author={Deng, Xinke and Mousavian, Arsalan and Xiang, Yu and Xia, Fei and Bretl, Timothy and Fox, Dieter},
  journal={IEEE Transactions on Robotics},
  volume={37},
  number={5},
  pages={1328--1342},
  year={2021},
  publisher={IEEE}
}

@inproceedings{li2018deepim,
  title={Deepim: Deep iterative matching for 6d pose estimation},
  author={Li, Yi and Wang, Gu and Ji, Xiangyang and Xiang, Yu and Fox, Dieter},
  booktitle={Proceedings of the European Conference on Computer Vision (ECCV)},
  pages={683--698},
  year={2018}
}

@inproceedings{zhao2023learning,
  title={Learning symmetry-aware geometry correspondences for 6d object pose estimation},
  author={Zhao, Heng and Wei, Shenxing and Shi, Dahu and Tan, Wenming and Li, Zheyang and Ren, Ye and Wei, Xing and Yang, Yi and Pu, Shiliang},
  booktitle={Proceedings of the IEEE/CVF International Conference on Computer Vision},
  pages={14045--14054},
  year={2023}
}

@inproceedings{chen2020g2l,
  title={G2l-net: Global to local network for real-time 6d pose estimation with embedding vector features},
  author={Chen, Wei and Jia, Xi and Chang, Hyung Jin and Duan, Jinming and Leonardis, Ales},
  booktitle={Proceedings of the IEEE/CVF conference on computer vision and pattern recognition},
  pages={4233--4242},
  year={2020}
}

@article{lin2024transpose,
  title={Transpose: 6d object pose estimation with geometry-aware transformer},
  author={Lin, Xiao and Wang, Deming and Zhou, Guangliang and Liu, Chengju and Chen, Qijun},
  journal={Neurocomputing},
  volume={589},
  pages={127652},
  year={2024},
  publisher={Elsevier}
}

@article{liu2025gnc,
  title={GNC-Pose: Geometry-Aware GNC-PnP for Accurate 6D Pose Estimation},
  author={Liu, Xiujin},
  journal={arXiv preprint arXiv:2512.06565},
  year={2025}
}

@article{lepetit2009ep,
  title={{EPnP}: An accurate {O(n)} solution to the {PnP} problem},
  author={Lepetit, Vincent and Moreno-Noguer, Francesc and Fua, Pascal},
  journal={International Journal of Computer Vision},
  volume={81},
  number={2},
  pages={155--166},
  year={2009},
  publisher={Springer}
}

@article{li2012robust,
  title={A robust O (n) solution to the perspective-n-point problem},
  author={Li, Shiqi and Xu, Chi and Xie, Ming},
  journal={IEEE Transactions on Pattern Analysis and Machine Intelligence},
  volume={34},
  number={7},
  pages={1444--1450},
  year={2012},
  publisher={IEEE}
}

@article{ulrich2011combining,
  title={Combining scale-space and similarity-based aspect graphs for fast 3D object recognition},
  author={Ulrich, Markus and Wiedemann, Christian and Steger, Carsten},
  journal={IEEE transactions on pattern analysis and machine intelligence},
  volume={34},
  number={10},
  pages={1902--1914},
  year={2011},
  publisher={IEEE}
}

@inproceedings{payet2011contours,
  title={From contours to 3d object detection and pose estimation},
  author={Payet, Nadia and Todorovic, Sinisa},
  booktitle={2011 International Conference on Computer Vision},
  pages={983--990},
  year={2011},
  organization={IEEE}
}

@inproceedings{wen2024foundationpose,
  title={Foundationpose: Unified 6d pose estimation and tracking of novel objects},
  author={Wen, Bowen and Yang, Wei and Kautz, Jan and Birchfield, Stan},
  booktitle={Proceedings of the IEEE/CVF Conference on Computer Vision and Pattern Recognition},
  pages={17868--17879},
  year={2024}
}

@article{yin2021graph,
  title={Graph neural network for 6D object pose estimation},
  author={Yin, Pengshuai and Ye, Jiayong and Lin, Guoshen and Wu, Qingyao},
  journal={Knowledge-Based Systems},
  volume={218},
  pages={106839},
  year={2021},
  publisher={Elsevier}
}

@inproceedings{lian2023checkerpose,
  title={Checkerpose: Progressive dense keypoint localization for object pose estimation with graph neural network},
  author={Lian, Ruyi and Ling, Haibin},
  booktitle={Proceedings of the IEEE/CVF International Conference on Computer Vision},
  pages={14022--14033},
  year={2023}
}

@inproceedings{rezazadeh2023hierarchical,
  title={Hierarchical Graph Neural Networks for Proprioceptive 6D Pose Estimation of In-hand Objects},
  author={Rezazadeh, Alireza and Dikhale, Snehal and Iba, Soshi and Jamali, Nawid},
  booktitle={2023 IEEE International Conference on Robotics and Automation (ICRA)},
  pages={2884--2890},
  year={2023}
}

@inproceedings{he2016deep,
  title={Deep residual learning for image recognition},
  author={He, Kaiming and Zhang, Xiangyu and Ren, Shaoqing and Sun, Jian},
  booktitle={Proceedings of the IEEE conference on computer vision and pattern recognition},
  pages={770--778},
  year={2016}
}

@article{vaswani2017attention,
  title={Attention is all you need},
  author={Vaswani, Ashish and Shazeer, Noam and Parmar, Niki and Uszkoreit, Jakob and Jones, Llion and Gomez, Aidan N and Kaiser, {\L}ukasz and Polosukhin, Illia},
  journal={Advances in neural information processing systems},
  volume={30},
  year={2017}
}

@inproceedings{martins2016softmax,
  title={From softmax to sparsemax: A sparse model of attention and multi-label classification},
  author={Martins, Andre and Astudillo, Ramon},
  booktitle={International conference on machine learning},
  pages={1614--1623},
  year={2016},
  organization={PMLR}
}

@book{press2007numerical,
  title={Numerical recipes 3rd edition: The art of scientific computing},
  author={Press, William H},
  year={2007},
  publisher={Cambridge university press}
}

@article{mccarthy1993generalized,
  title={Generalized legendre polynomials},
  author={McCarthy, PC and Sayre, JE and Shawyer, BLR},
  journal={Journal of mathematical analysis and applications},
  volume={177},
  number={2},
  pages={530--537},
  year={1993},
  publisher={Elsevier}
}

\end{document}